\documentclass{article}

\usepackage[preprint]{neurips_2024}

\usepackage[utf8]{inputenc} 
\usepackage[T1]{fontenc}    
\usepackage{hyperref}       
\usepackage{url}            
\usepackage{booktabs}       
\usepackage{amsfonts}       
\usepackage{nicefrac}       
\usepackage{microtype}      
\usepackage{xcolor}         
\usepackage{multirow}
\usepackage{graphicx}

\usepackage{natbib}
\setcitestyle{numbers,square}

\title{Explainable Behavior Cloning: Teaching Large Language Model Agents through Learning by Demonstration}

\author{
\textbf{Yanchu Guan$^{1}$, Dong Wang$^{1}$, Yan Wang$^{1}$, Haiqing Wang$^{1}$, Renen Sun$^{1}$,} \\
\textbf{Chenyi Zhuang$^{1}$, Jinjie Gu$^{1}$, Zhixuan Chu$^{2\spadesuit}$} \\
   $^1$ Ant Group\\
  $^2$ Zhejiang University\\
  Hangzhou, China \\
  \texttt{\{yanchu.gyc,yishan.wd,luli.wy\}@antgroup.com, zhixuanchu@zju.edu.cn} \\
}

\begin{document}

\def\thefootnote{$\spadesuit$}\footnotetext{Corresponding author.}
\def\thefootnote{\arabic{footnote}}

\maketitle

\begin{abstract}
Autonomous mobile app interaction has become increasingly important with growing complexity of mobile applications. Developing intelligent agents that can effectively navigate and interact with mobile apps remains a significant challenge. In this paper, we propose an Explainable Behavior Cloning LLM Agent (EBC-LLMAgent), a novel approach that combines large language models (LLMs) with behavior cloning by learning demonstrations to create intelligent and explainable agents for autonomous mobile
app interaction. EBC-LLMAgent consists of three core modules: Demonstration Encoding, Code Generation, and UI Mapping, which work synergistically to capture user
demonstrations, generate executable codes, and establish accurate correspondence between code and UI elements. We introduce the Behavior Cloning Chain Fusion technique to enhance the generalization capabilities of the agent. Extensive experiments on five popular mobile applications from diverse domains demonstrate the superior performance of EBC-LLMAgent, achieving high success rates in task completion, efficient generalization to unseen scenarios, and the generation of meaningful explanations.
\end{abstract}

\section{Introduction}
Mobile applications have become ubiquitous in our daily lives, offering a wide range of functionalities and services. With the increasing complexity and diversity of mobile apps, there is a growing need for intelligent agents that can autonomously interact with these apps, assisting users in various tasks and enhancing their overall experience. Autonomous mobile app interaction involves understanding the app's user interface, executing appropriate actions, and providing transparent explanations of the agent's behavior. However, developing such agents poses significant challenges due to the diverse and dynamic nature of mobile app interfaces and the need for interpretable and generalizable interaction strategies. Traditional approaches to mobile app automation often rely on hand-crafted rules and heuristics, which are labor-intensive to create and maintain, and struggle to adapt to new scenarios and app updates. Recent advancements in large language models (LLMs) \cite{gpt3-conf/nips/BrownMRSKDNSSAA20, llama-journals/corr/abs-2302-13971, InstructGPT-conf/nips/Ouyang0JAWMZASR22} have shown remarkable success in natural language understanding and generation tasks, demonstrating their ability to capture and leverage vast amounts of knowledge from diverse sources \cite{5-journals/tmlr/WeiTBRZBYBZMCHVLDF22, 6-conf/nips/SchaefferMK23, 7-journals/corr/abs-2001-08361}. However, the application of LLMs in the context of autonomous mobile app interaction remains largely unexplored\cite{AppAgent-abs-2312-13771}.

In this paper, we propose a novel approach called Explainable Behavior Cloning LLM Agent (EBC-LLMAgent) that combines the power of LLMs with behavior cloning by learning demonstrations to create intelligent and explainable agents for autonomous mobile app interaction. Our approach aims to address the limitations of traditional methods by leveraging the generalization capabilities of LLMs and incorporating techniques for explainable and transparent decision-making. The main motivation behind our work is to develop intelligent agents that can autonomously navigate and interact with mobile apps, reducing the need for manual intervention and enhancing user productivity. By learning from user demonstrations, our approach enables agents to capture and replicate complex interaction patterns, adapting to different app layouts and functionalities. Moreover, by providing transparent explanations of the agent's actions, our approach aims to build user trust and facilitate seamless human-agent collaboration.

The key novelty of our approach lies in the integration of LLMs with behavior cloning by learning demonstrations, enabling the generation of executable code snippets that replicate demonstrated behaviors. We propose a modular architecture consisting of three core components: Demonstration Encoding, Code Generation, and UI Mapping. The Demonstration Encoding module captures and structures user demonstrations into a format processable by the LLM agent, leveraging advanced visual question answering models to extract rich semantic information. The Code Generation module utilizes the generative capabilities of LLMs to translate encoded demonstrations into modular, parameterized, and explanatory code snippets. The UI Mapping module establishes a correspondence between the generated code snippets and the relevant UI elements within the app, ensuring accurate and seamless interaction. Furthermore, we introduce the Behavior Cloning Chain Fusion technique, which allows the agent to learn from multiple demonstrations and merge the learned behaviors into a cohesive and flexible interaction model. This technique enhances the generalization capabilities of the agent by enabling it to adapt to new scenarios efficiently, combining and executing appropriate learned functions based on the recognized task requirements.

\begin{figure*}[t]
    \centering
    \includegraphics[width=1\columnwidth]{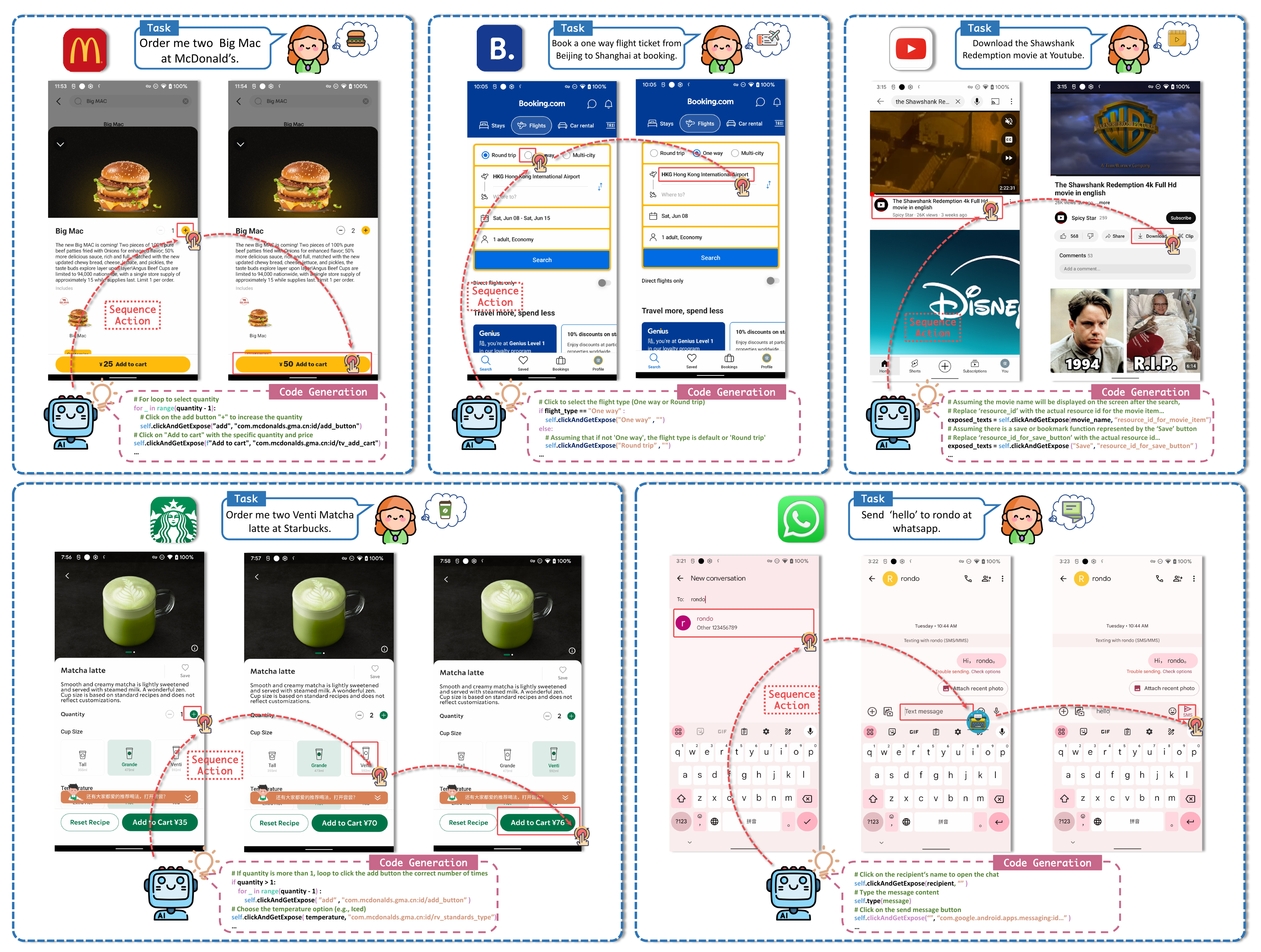}
    
    \caption{Real-world examples of Code Generation in EBC-LLMAgent across various mobile applications. Each represents a different task: (a) Ordering from McDonald's, (b) Booking a flight ticket, (c) Downloading a movie on YouTube, (d) Ordering from Starbucks, and (e) Sending a message on WhatsApp. These examples showcase the agent's ability to handle diverse tasks across different app interfaces, generating modular and interpretable code that bridges the gap between user intent and app-specific actions.}
    
    \label{fig:real_case}
\end{figure*}

To validate the effectiveness and practicality of our approach, we conduct extensive experiments on five popular mobile applications from diverse domains, including dining, entertainment, travel, and communication. Figure ~\ref{fig:real_case} shows the examples of Code Generation in EBC-LLMAgent across various mobile applications. The experimental results demonstrate the superior performance of EBC-LLMAgent compared to baseline methods, achieving high success rates in task completion, efficient generalization to unseen scenarios, and the generation of meaningful explanations. The videos of Demonstration Encoding and UI Mapping are provided in the supplementary materials, which showcase the process of behavior cloning learning from user demonstrations and the system automatically executing the steps within the app.

The main contributions of this paper are as follows:

\begin{itemize}
  \item We propose EBC-LLMAgent, a novel approach that combines LLMs with behavior cloning by learning demonstrations for autonomous mobile app interaction, enabling intelligent and explainable agents to learn from user demonstrations and generalize to unseen tasks.

  \item We introduce a modular architecture consisting of Demonstration Encoding, Code Generation, and UI Mapping modules, which work synergistically to capture user demonstrations, generate executable code snippets, and establish accurate correspondence between code and UI elements.

  \item We propose the Behavior Cloning Chain Fusion technique, which enhances the generalization capabilities of the agent by enabling it to learn from multiple demonstrations and merge learned behaviors into a cohesive and flexible interaction model.
\end{itemize}

\section{Related Work}

Our work builds upon and integrates techniques from several areas, including robotic process automation, web navigation, and code generation. We provide an overview of relevant prior work in each of these domains and discuss how our proposed approach advances the state-of-the-art.

\subsection{Robotic Process Automation}

Robotic Process Automation (RPA) focuses on automating repetitive and rule-based tasks traditionally performed by humans interacting with software applications \cite{14-hofmann2020robotic, 15-ivanvcic2019robotic, 16-syed2020robotic}. RPA tools aim to replicate human actions, such as clicking buttons, entering data, and navigating interfaces, to streamline business processes and improve efficiency. Existing RPA approaches often rely on hard-coded rules and heuristics to define automation workflows \cite{17-seguin2021minimizing, 18-conf/kes/TremblaySBPL23}. These approaches require significant manual effort to create and maintain, and struggle to adapt to changes in application interfaces or handle exceptional cases. More recently, there has been growing interest in integrating AI and machine learning techniques into RPA to enable more intelligent and adaptable automation \cite{19-journals/corr/abs-2311-10751, 20-conf/icpm/SaniSB23}. However, current AI-powered RPA solutions still face challenges in generalizing to new tasks, providing transparent explanations of their actions, and leveraging the vast knowledge captured in large language models. Our work addresses these limitations by combining the principles of RPA with the power of LLMs and learning by demonstration techniques.
We aspire to create a more robust automation framework that not only streamlines repetitive processes but also intelligently adapts to novel tasks, thus enhancing overall efficiency and effectiveness.  Additionally, this integration seeks to provide clearer explanations of automated actions, fostering greater user trust and enabling organizations to maximize the potential of their automated systems in a rapidly evolving digital environment.

\subsection{Web Navigation}

Autonomous web navigation is a closely related field that aims to develop agents capable of browsing and interacting with websites to accomplish specific goals \cite{21-conf/sii/SuenagaM20, 22-journals/ijpcc/TanwarSM23, guan2023intelligent, deng2024mind2web, gur2023real, rawles2023android, zhan2023you}. Early approaches relied on manually defined rules and heuristics to guide the navigation process. More recently, there has been a shift towards data-driven methods that learn navigation policies from demonstrations or through reinforcement learning. One notable line of work focuses on using natural language instructions to guide web navigation. These approaches aim to map high-level instructions to low-level actions required to navigate and interact with web pages. They are mainly divided into two categories. The first category consists of single modal models that use HTML for element recall to construct a candidate set, thereby reducing the search space and allowing the model to make selections from it. The second category consists of end-to-end multimodal models that directly learn the operation types and the coordinates of the elements. In the case of multi-step tasks, a history is constructed in the prompt to assist the model in reasoning. However, they often struggle with complex or unseen instructions and lack the ability to provide clear explanations of their decision-making process. As the number of steps in the task increases, the history will become much longer, significantly increasing the difficulty of training the model. Our approach extends the ideas from web navigation to the domain of mobile app interaction. We leverage the power of LLMs to enable natural language understanding and generation, allowing our agent to interpret high-level instructions and provide transparent explanations of its actions.

\subsection{Code Generation}

Code generation techniques aim to automatically produce executable code based on various inputs \cite{23-conf/acl/0002LJ23, 24-conf/icml/ZhengSJWXXW23, 25-conf/icml/ZhangYHLYF023}, such as natural language descriptions, examples, or partial code snippets. Recent advancements in large language models, such as GPT-3 and Codex, have shown impressive capabilities in generating code across multiple programming languages. One relevant application of code generation is in the context of task automation. For example, \cite{agostinelli2020automated} proposed a method for generating executable scripts from natural language descriptions to automate web browsing tasks. \cite{hu2024deploying} used a combination of LLMs and program synthesis techniques to generate code for robotic task planning and execution. However, existing code generation approaches often struggle with generating code that accurately maps to specific UI elements and handles the dynamic nature of mobile app interfaces. Our approach addresses these challenges by integrating demonstration encoding, UI mapping, and behavior cloning techniques to enable the generation of executable code snippets that replicate demonstrated app interactions. In the process of automating tasks, coding is widely regarded as a highly efficient and stable solution. Code inherently supports branching and looping logic, making it an ideal tool for tackling complex tasks. By effectively extracting and managing parameters, code can adapt flexibly to unseen tasks, showing excellent generalization capabilities. Moreover, through diverse demonstrations, code can operate efficiently in various scenarios, comprehensively covering all aspects of complex workflows. In summary, the flexibility and universality of code provide a solid foundation for the effective execution of automated tasks.

\begin{figure*}[t]
    \centering
    \includegraphics[width=1\columnwidth]{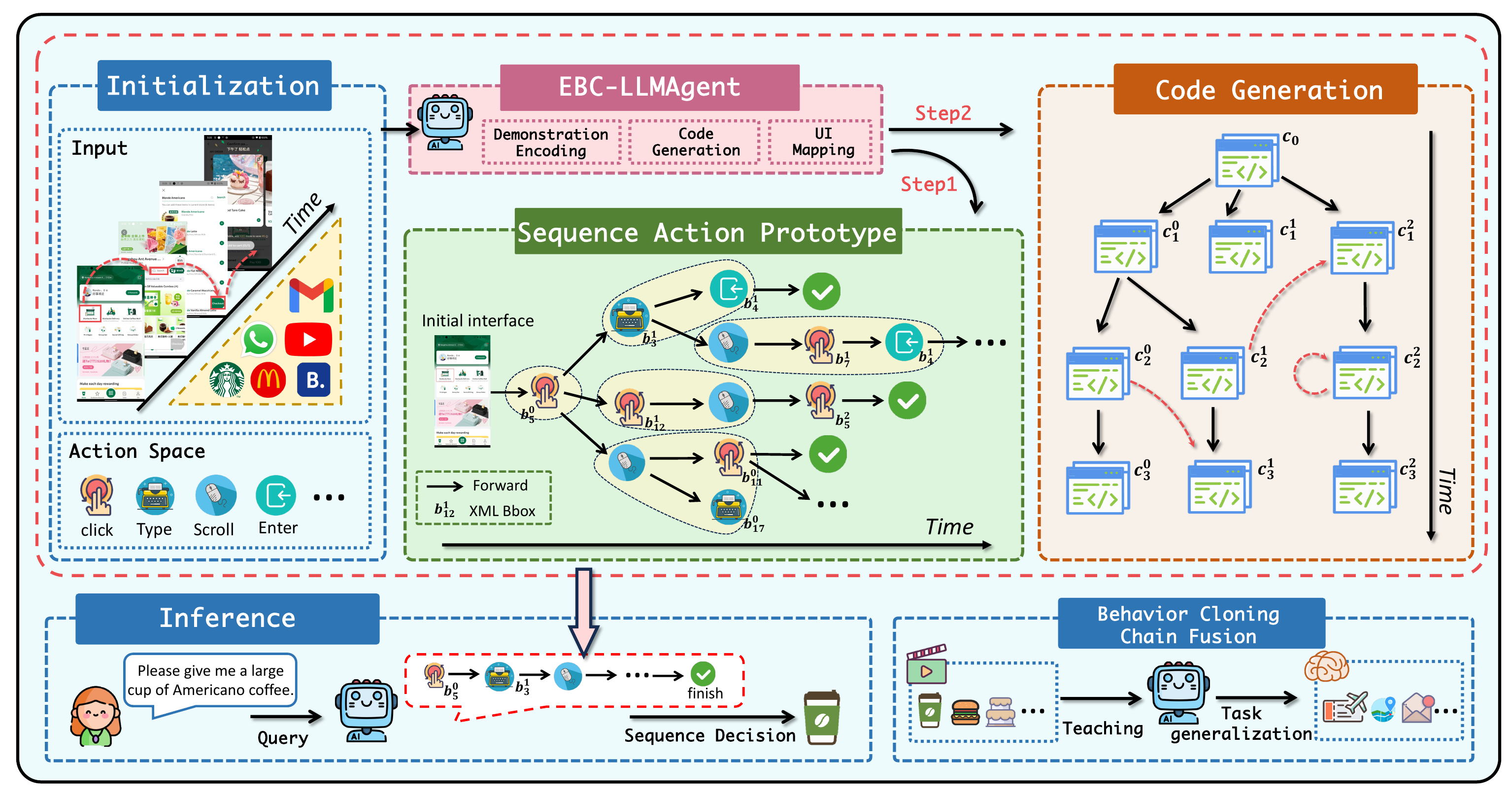}
    
    \caption{The framework of our proposed EBC-LLMAgent.}
    
    \label{fig:framework}
\end{figure*}

\section{Methodology}
As shown in Figure \ref{fig:framework}, we present a novel approach that combines large language models (LLMs) with behavior cloning by learning demonstrations to create explainable agents for autonomous mobile app interaction. Our methodology, named Explainable Behavior Cloning LLM Agent (EBC-LLMAgent), consists of three core modules: Demonstration Encoding, Code Generation, and UI Mapping. These modules work synergistically to enable LLM agents to learn from user demonstrations, generalize to unseen tasks, and provide transparent explanations of their actions.

\subsection{Demonstration Encoding}
The Demonstration Encoding module captures and structures user demonstrations into a format processable by the LLM agent. A user demonstration is represented as a sequence of actions performed within the mobile app, denoted as $D = \{a_1, a_2, \ldots, a_n\}$, where each action $a_i$ is a tuple $(\tau_i, e_i, m_i)$ consisting of the action type $\tau_i$ (e.g., click, type, scroll, enter, and back), the interacted element $e_i$, and the associated metadata $m_i$ (e.g., text, identifier, bounds). It is important to note that elements and their corresponding metadata can be retrieved from the page content, such as XML or DOM. Associating this with the action enables the possibility of reproducing the operation.

The encoding process transforms the demonstration $D$ into a structured representation $E_D = \{s_1, s_2, \ldots, s_n\}$, where each encoded step $s_i$ is a tuple $(\tau_i, t_i, id_i, v_i, exp_i)$ containing the action type $\tau_i$, the associated text $t_i$, the identifier $id_i$ of the interacted element (which may not be unique or exist), the extracted visual features $v_i$ obtained using models like Q-wen VL or GPT-4v for visual question answering (VQA) referring tasks, and the list of exposed texts $exp_i$ on the screen.

The visual features $v_i$ play a crucial role in enabling the agent to understand and interact with the app's user interface. It is a text representation based on regional image, when $t_i$ and $id_i$ fail to identify the target element, $v_i$ can assist in achieving unique identification of the element. By leveraging advanced VQA models, the Demonstration Encoding module captures rich semantic information about the interacted elements, allowing the agent to generalize to unseen scenarios and provide accurate explanations of its actions.

\subsection{Code Generation}
The Code Generation module leverages the generative capabilities of the LLM to translate the encoded demonstration $E_D$ into executable code. The LLM, denoted as $\mathcal{L}$, is prompted with the encoded steps $s_i$ along with the app metadata $M$ and generates code snippets $C = \{c_1, c_2, \ldots, c_n\}$ that replicate the demonstrated behavior. The generation process can be formulated as:
$c_i = \mathcal{L}(s_i, M, \theta)$
where $\theta$ represents the learnable parameters of the LLM. 

It's important to note that the number of generated code snippets may be less than the number of demonstration steps ($|C| \leq |E_D|$) due to potential loop structures in the code. The Code Generation module intelligently identifies and leverages these loop structures to generate concise and efficient code that accurately replicates the demonstrated behavior.

To enable generalization to unseen actions, the Code Generation module identifies and extracts relevant hyperparameters $H = \{h_1, h_2, \ldots, h_k\}$ from the demonstration using image recognition techniques. These hyperparameters form a set of choices that allow the agent to adapt the generated code dynamically based on the recognized parameters. The extraction of hyperparameters can be represented as:
$h_j = \mathcal{R}(v_i, \phi)$
where $\mathcal{R}$ is an image recognition model with learnable parameters $\phi$.

The generated code snippets $c_i$ are designed to be modular, parameterized, and accompanied by explanatory comments to ensure transparency and interoperability. This allows for easy understanding and modification of the generated code, enhancing the explainability and adaptability of the EBC-LLMAgent. With this approach, you only need to demonstrate the process of ordering an Americano once, and our EBC-LLMAgent can learn how to order a latte, even helping you order 10 or 20 cups. This is unimaginable for single step decision making models in web navigation area.

\subsection{UI Mapping}
The UI Mapping module establishes a correspondence between the generated code snippets $c_i$ and the relevant UI elements within the app. It employs two main approaches: text and identifier matching, and visual information matching. The text and identifier matching approach locates the target element for interaction by matching the text and resource identifier of UI elements. This approach leverages the structured representation of the app's UI hierarchy to find the most relevant element based on the encoded step information. On the other hand, the visual information matching approach utilizes visual features and grounding algorithms to locate UI elements based on their appearance. By comparing the extracted visual features $v_i$ with the visual information of the UI elements, the UI Mapping module can accurately identify the target element even in cases where the text or identifier information is ambiguous or missing.

The UI Mapping module can be formulated as:
$u_i = \mathcal{M}(s_i, U, \psi)$
where $u_i$ is the corresponding UI element for the encoded step $s_i$, $U$ is the app's UI hierarchy, and $\psi$ represents the learnable parameters of the mapping function $\mathcal{M}$. The UI Mapping module ensures that the generated code can be executed seamlessly within the app, mimicking the user's actions with high fidelity. It also provides explanations of how the agent identifies and interacts with specific UI elements, enhancing the transparency and interpretability of the agent's behavior.

\subsection{Behavior Cloning Chain Fusion}
To further enhance the generalization capabilities of our approach, we introduce the Behavior Cloning Chain Fusion technique. This technique allows the agent to learn from multiple demonstrations and merge the learned behaviors into a cohesive and flexible interaction model.

Let $\mathcal{D} = \{D_1, D_2, \ldots, D_m\}$ be a set of $m$ demonstrations teaching different tasks. Each demonstration $D_i$ is encoded and processed through the Code Generation module, resulting in a set of learned behaviors represented as code functions $\mathcal{F} = \{f_1, f_2, \ldots, f_m\}$. The Behavior Cloning Chain Fusion module, denoted as $\mathcal{B}$, dynamically invokes and combines the learned functions based on the recognized task requirements. Given a new task $T$, the fusion process can be formulated as:
$\hat{f} = \mathcal{B}(T, \mathcal{F}, \xi)$
where $\hat{f}$ is the fused behavior function and $\xi$ represents the learnable parameters of the fusion module. The fused behavior function $\hat{f}$ intelligently selects and executes the appropriate learned functions, enabling the agent to adapt to new scenarios efficiently. By leveraging the knowledge gained from multiple demonstrations, the agent can handle a wider range of tasks and exhibit more robust and flexible behavior.

The Behavior Cloning Chain Fusion module plays a crucial role in enhancing the generalization ability of the EBC-LLMAgent. It allows the agent to combine learned behaviors in novel ways, enabling it to tackle unseen tasks by composing and adapting the knowledge acquired from different demonstrations.

\section{Experiments}

We evaluate the proposed EBC-LLMAgent through various experiments to demonstrate its effectiveness. Before presenting the experimental results, we first describe the experimental setup, including the data used and the evaluation metrics. The videos of Demonstration Encoding and UI Mapping are provided in the supplementary materials, which showcase the process of behavior cloning learning from user demonstrations and the system automatically executing the steps within the app.

\subsection{Experimental Setup}
\textbf{Dataset.} We conducted extensive experiments on five popular applications: Starbucks, McDonald's, Booking, YouTube, and WhatsApp, to validate the practicality of EBC-LLMAgent. These applications represent diverse fields such as dining, entertainment, travel, and communication, offering a wide range of testing scenarios. Among them, tasks for Starbucks, McDonald's, and Booking are relatively complex, averaging more than nine steps. Meanwhile, tasks for YouTube and WhatsApp are comparatively simple, with an average of around five steps each. We designed more than thirty different tasks for each application. For example, the Starbucks ordering task encompassed various parameters such as dish names, quantities, specifications, and pickup options. To ensure the fairness of our evaluation, we performed at least ten trials for each task, using the average value as the metric.

\textbf{Environment.} All experiments testing
run on the Linux server(Ubuntu 16.04) with the Intel(R) Xeon(R)
Silver 4214 2.20GHz CPU, 512GB memory, and 1 NVIDIA A-100
GPU.

\textbf{Evaluation Metrics.} We employed three distinct metrics for analysis:
\begin{itemize}
    \item \textbf{Task Completion Rate (Task CR)}: This metric measures the progress of task completion, calculated by dividing the number of steps successfully completed ($finished\_steps$) by the total number of steps required ($total\_steps$) for the task.
    \item \textbf{Task Success Rate (Task SR)}: A task is considered successful when all of its constituent steps have been completed successfully. 
    \item \textbf{Average Steps (Avg Steps)}: This metric determines the average number of steps taken to successfully complete a task.
\end{itemize}

\subsection{Experimental Results}
In this part, the experiments are designed to answer the following questions:
\begin{itemize}
    \item \textbf{Q1}: Does our proposed EBC-LLMAgent outperform other approaches in the field of Autonomous Mobile App Interaction?
    \item \textbf{Q2}: How does each variant of the proposed EBC-LLMAgent enhance overall performance?
    \item \textbf{Q3}: How resilient is the EBC-LLMAgent to changes in experimental settings?
    \item \textbf{Q4}: What is the generalization capability of EBC-LLMAgent?
    \item \textbf{Q5}: What is the explainability capability of EBC-LLMAgent?
\end{itemize}

\begin{table}[h]
\centering
\caption{Experimental results on five popular mobile applications. To ensure fairness in the comparison, we employed a uniform action space(click, type, scroll, enter, and back). GPT-4v was used as our baseline. On this basis, In-Context Learning(ICL) and React enhance performance by incorporating additional contextual information and multiple rounds of inquiries. AppAgent integrates extra documentation annotations for certain tasks. Moreover, Task CR stands for Task Completion Rate, Task SR stands for Task Success Rate, and Avg Steps stands for Average Steps.}
\label{table:main_resuls}
\begin{tabular}{l|c|ccccc}
\toprule
AppName & Indicator & GPT-4v & ICL & React & AppAgent & Ours \\
\midrule
\multirow{3}{*}{McDonald's} & Task CR & 32.7 & 37.6 & 43.9 & 66.4 & \textbf{95.6} \\
 & Task SR & 0 & 0 & 0 & 0 & \textbf{91.7} \\
 & Avg Steps & 0 & 0 & 0 & 0 & 8.9 \\
\midrule
\multirow{3}{*}{Starbucks} & Task CR & 35.0 & 38.1 & 45.8 & 65.2 & \textbf{96.0} \\
 & Task SR & 0 & 0 & 0 & 0 & \textbf{92.4} \\
 & Avg Steps & 0 & 0 & 0 & 0 & 8.6 \\
\midrule
\multirow{3}{*}{Booking} & Task CR & 14.9 & 25.4 & 33.7 & 37.2 & \textbf{93.1} \\
 & Task SR & 0 & 0 & 0 & 0 & \textbf{90.3} \\
 & Avg Steps & 0 & 0 & 0 & 0 & 13.5 \\
\midrule
\multirow{3}{*}{YouTube} & Task CR & 70.8 & 74.2 & 77.5 & 88.2 & \textbf{96.7} \\
 & Task SR & 49.2 & 51.3 & 55.1 & 85.3 & \textbf{94.2} \\
 & Avg Steps & 6.1 & 6.0 & 7.2 & 5.5 & 5.2 \\
\midrule
\multirow{3}{*}{WhatsApp} & Task CR & 64.5 & 67.7 & 72.6 & 87.3 & \textbf{96.6} \\
 & Task SR & 47.9 & 50.4 & 53.2 & 84.8 & \textbf{94.7} \\
 & Avg Steps & 6.2 & 6.3 & 7.3 & 5.7 & 5.1 \\
\bottomrule
\end{tabular}
\end{table}

\subsection*{ Q1: Effectiveness of EBC-LLMAgent}

To holistically assess the performance of our proposed EBC-LLMAgent, , we conducted a comprehensive comparison with GPT-4v, In-Context Learning\cite{gpt3-conf/nips/BrownMRSKDNSSAA20}, React\cite{react-conf/icml/ZhangYHLYF023}, and AppAgent\cite{AppAgent-abs-2312-13771}.

Table~\ref{table:main_resuls} presents the principal experimental outcomes of different models on five applications. For tasks with fewer steps, such as YouTube and WhatsApp, GPT-4v attains a Task Success Rate of about 50\%, while AppAgent is close to 85\% and EBC-LLMAgent is nearly 94\%. For tasks entailing a greater number of steps like Booking, Starbucks, and McDonald's, these models commonly face difficulties in completion. However, our EBC-LLMAgent can complete these tasks with high success rate. Regarding Task Completion Rate, as the number of steps increases, the GPT-4v sharply drops from 70\% to 32\%, whereas AppAgent's performance declines from 88\% to 65\%. Our EBC-LLMAgent consistently achieves optimal performance in all five applications, with a Task Success Rate exceeding 90\%.

Our experimental results demonstrate that the EBC-LLMAgent is a truly practical agent capable of reliably completing complex tasks and exhibiting outstanding performance across various scenarios. Compared to several similar studies, our agent shows significant advantages in problem-solving, particularly in terms of success rates. While other research has nearly zero success rates when facing slightly more complex tasks, the EBC-LLMAgent consistently maintains strong performance.

This stability and high success rate not only highlight our superiority in algorithm design and implementation but also provide robust support for practical applications. Whether in industrial production, medical diagnosis, or intelligent customer service, the EBC-LLMAgent effectively addresses the ever-changing demands and challenges. This validates the effectiveness of our research and lays a solid foundation for tackling more complex tasks in the future. The application potential of EBC-LLMAgent in the field of artificial intelligence is immense and holds promise for driving more real-world applications.

\begin{table}[h]
\centering

\caption{Ablation experimental results. Our ablation experiments concentrate on the Demonstration Encoding and UI Mapping modules. More precisely, Surrounding Features refers to the application of adjacent contextual data for UI element mapping, w/o Surrounding\&Visual Features denotes a reliance exclusively on the element's inherent text and identifier.}

\label{table:ablation_study}
\resizebox{1\columnwidth}{!}{
\begin{tabular}{l|c|cccc}
\toprule
AppName & Indicator & GPT-4v & w/o Surrounding\&Visual Features & w/o Visual Features & Ours \\
\midrule
\multirow{3}{*}{McDonald's} & Task CR & 32.7 & 78.0 & 91.2 & \textbf{95.6} \\
 & Task SR & 0 & 73.4 & 87.5 & 91.7 \\
 & Avg Steps & 0 & 7.5 & 8.3 & 8.9 \\
\midrule
\multirow{3}{*}{Starbucks} & Task CR & 35.0 & 74.5 & 91.3 & \textbf{96.0} \\
 & Task SR & 0 & 71.7 & 87.9 & \textbf{92.4} \\
 & Avg Steps & 0 & 7.4 & 8.2 & 8.6 \\
\midrule
\multirow{3}{*}{Booking} & Task CR & 14.9 & 82.7 & 91.2 & \textbf{93.1} \\
 & Task SR & 0 & 83.2 & 89.8 & \textbf{90.3} \\
 & Avg Steps & 0 & 12.7 & 13.2 & 13.5 \\
\bottomrule
\end{tabular}}
\end{table}

\subsection*{ Q2: Ablation Study}

To thoroughly evaluate the impact of various variants in our proposed EBC-LLMAgent, we conducted several ablation studies. Surrounding Features refer to the information surrounding the elements, while Visual Features pertain to the textual descriptions extracted by the multimodal model.

As shown in Table~\ref{table:ablation_study}, the version without Surrounding\&Visual Features performs poorly as expected. In fact, the impact of surrounding features on the metrics is pronounced. In app navigation, encountering text and styles that are identical is frequent. For example, in the Starbucks menu selection interface, each item might have an identical 'add' button. Merely recording information associated with this 'add' button can lead to imprecise targeting during UI Mapping, resulting in task failure. Visual features are also vital, it offers valuable support when elements cannot be uniquely located. By integrating these two designs, our EBC-LLMAgent can deliver excellent results.

\begin{table}[h]
\centering

\caption{Experimental results of using various LLMs. }

\label{table:different_basemodels}
\resizebox{1\columnwidth}{!}{
\begin{tabular}{l|ccccccc}
\toprule
AppName & Indicator & llama2-13b & Vicuna-13b & Baichuan2-13B & Qwen-14B & GPT-3.5 & GPT-4 \\
\midrule
\multirow{3}{*}{McDonald's} & Task SR & 43.1 & 51.4 & 49.5 & 70.6 & 82.5 & \textbf{91.7} \\
 & Avg Steps & 7.1 & 7.2 & 7.4 & 7.3 & 8.7 & 8.9 \\
\midrule
\multirow{3}{*}{Starbucks} & Task SR & 43.5 & 51.6 & 50.7 & 72.1 & 81.3 & \textbf{92.4} \\
 & Avg Steps & 7.2 & 7.3 & 7.1 & 7.2 & 8.8 & 8.6 \\
\midrule
\multirow{3}{*}{Booking} & Task SR & 42.3 & 50.9 & 49.7 & 69.3 & 80.2 & \textbf{90.3} \\
 & Avg Steps & 11.4 & 11.5 & 11.8 & 11.4 & 13.1 & 13.5 \\
\bottomrule
\end{tabular}}
\end{table}

\subsection*{ Q3: Sensitivity Analysis}

We will conduct the analysis from the following three dimensions.

\paragraph{a. Different Code Generation Models} 

We fix the Demonstration Encoding and UI Mapping within the framework, utilizing different LLMs for the Code Generation component. As indicated in Table~\ref{table:different_basemodels}, among models of similar sizes, Qwen-14b achieved the best performance. Among all evaluated models, GPT-4 achieves the highest overall results. This emphasizes the importance of selecting the right model based on specific criteria, such as size and performance metrics, to achieve optimal outcomes in code generation. Through this comprehensive evaluation, we aim to leverage the strengths of these advanced LLMs, ensuring effective and efficient code generation that aligns with our framework's objectives.

\begin{table}[h]
\centering

\caption{Experimental results with different scaling models.}

\label{table:scaling_law}
\begin{tabular}{l|cccc}
\toprule
AppName & Indicator & Vicuna-7b & Vicuna-13b & Vicuna-33b \\
\midrule
\multirow{3}{*}{McDonald's} & Task SR & 42.3 & 51.4 & \textbf{53.5} \\
 & Avg Steps & 7.4 & 7.1 & 7.3 \\
\midrule
\multirow{3}{*}{Starbucks} & Task SR & 43.5 & 51.6 & \textbf{54.0} \\
 & Avg Steps & 7.2 & 7.3 & 7.2 \\
\midrule
\multirow{3}{*}{Booking} & Task SR & 43.0 & 50.9 & \textbf{52.9} \\
 & Avg Steps & 11.4 & 11.5 & 11.8 \\
\bottomrule
\end{tabular}
\end{table}

\paragraph{b. Scaling Law}

We investigated the impact of the size of our Code Generation Model on its performance. As presented in Table~\ref{table:scaling_law}, we employed models from the Vicuna series with 7b, 13b, and 33b parameters, respectively. It was observed that the 13b model demonstrated approximately a 20\% improvement in performance over the 7b model across three different apps. However, when increasing the model size from 13b to 33b, there is only about a 4\% enhancement observed. 
This indicates that as the model size increases, the returns will grow, but the marginal returns will gradually diminish. Therefore, we need to tradeoff the model's size against its inference performance to choose the appropriate version.

\begin{figure*}[t!]
    \centering
    \includegraphics[width=1\columnwidth]{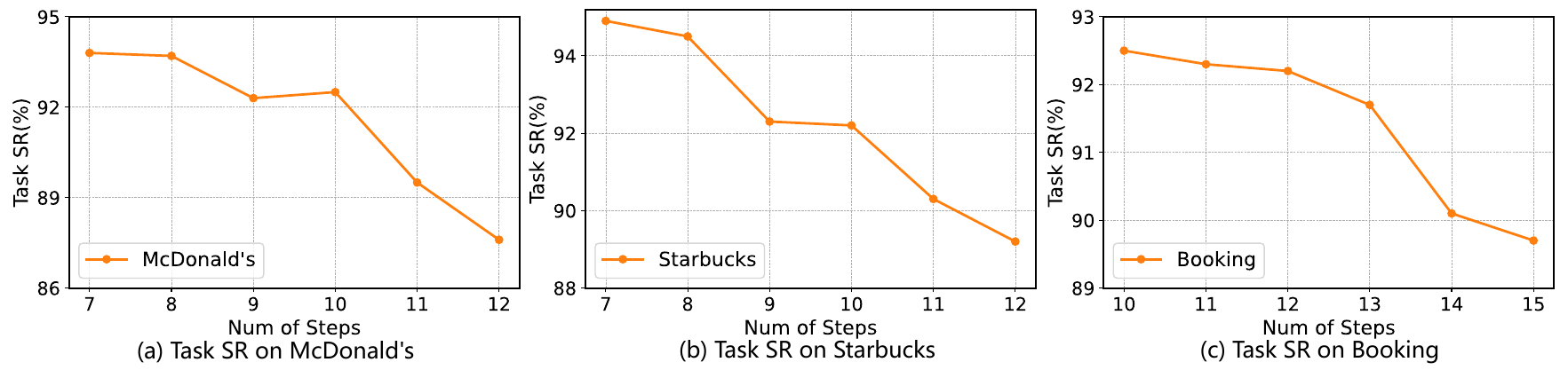}

    \caption{Task SR tends to decrease as the number of steps increases.}

    \label{figure:task_steps}
\end{figure*}

\paragraph{c. Steps for Task SR}

As shown in Figure~\ref{figure:task_steps}, Task SR tends to decrease as the number of steps increases. This trend aligns with our expectations, as tasks with more steps typically correspond to higher complexity. However, even when task sequences extend to 12 steps, our framework still maintains a success rate of over 87\% on McDonald's and Starbucks. However, even when the task reaches 12 steps, the success rate still exceeds 92\% on Booking. This demonstrates our framework's capability to tackle a wide range of complex tasks on real-world mobile applications.

\begin{table}[h]
\centering
\caption{Cross Type experimental results.}
\begin{tabular}{l|c|c|ccc}
\toprule
AppName & Cross Type & Indicator & GPT-4v  & AppAgent & Ours \\
\midrule
\multirow{3}{*}{McDonald's} & Combo2Hamburger & \multirow{3}{*}{Task CR} & 33.5 & 61.7 & \textbf{93.7} \\
 & Combo2Drinks &  & 35.1 & 64.2 & \textbf{95.3} \\
 & Hamburger2Drinks &  & 39.4 & 62.9 & \textbf{94.8} \\
\midrule
\multirow{3}{*}{Starbucks} & Combo2Coffee & \multirow{3}{*}{Task CR} & 37.1 & 58.3 & \textbf{91.5} \\
 & Combo2Dessert &  & 39.2 & 62.7 & \textbf{93.1} \\
 & Coffee2Dessert &  & 41.3 & 61.0 & \textbf{92.6} \\
\bottomrule
\end{tabular}
 \label{tab:cross_type}
\end{table}

\begin{table}[h]
\centering
\caption{Cross Scene experimental results.}
\begin{tabular}{l|c|c|ccc}
\toprule
AppName & Generalization & Indicator & GPT-4v & AppAgent & Ours \\
\midrule
McDonald's & Cross Scene & Task CR & 31.5 & 59.1 & \textbf{87.1} \\
Starbucks & Cross Scene & Task CR & 36.2 & 60.6 & \textbf{89.3} \\
\bottomrule
\end{tabular}
 \label{tab:cross_scene}
\end{table}

\subsection*{Q4: Generalization Analysis}

To assess the generalization capability of our proposed EBC-LLMAgent, we conducted cross-type and cross-scene experiments on the McDonald's and Starbucks applications. The results are presented in Table \ref{tab:cross_type} and Table \ref{tab:cross_scene}, respectively.

In the cross-type experiments (Table \ref{tab:cross_type}), we evaluated the agent's ability to generalize across different types of tasks within the same application. For McDonald's, we tested the agent's performance on tasks involving transitions from combo meals to hamburgers (Combo2Hamburger), combo meals to drinks (Combo2Drinks), and hamburgers to drinks (Hamburger2Drinks). Similarly, for Starbucks, we assessed the agent's generalization across tasks involving transitions from combo items to coffee (Combo2Coffee), combo items to desserts (Combo2Dessert), and coffee to desserts (Coffee2Dessert). We report the Task Completion Rate (Task CR) as the evaluation metric.
The results demonstrate that our EBC-LLMAgent significantly outperforms the GPT-4 baseline and AppAgent across all cross-type tasks. 

The cross-scene experiments (Table \ref{tab:cross_scene}) assess the agent's ability to generalize to different scenes or contexts within the same application. We evaluate the Task Completion Rate (Task CR) for both McDonald's and Starbucks applications. Our EBC-LLMAgent achieves impressive Task CR values of $87.1\%$ and $89.3\%$ for McDonald's and Starbucks, respectively, demonstrating its strong generalization capability across different scenes. In contrast, GPT-4 and AppAgent exhibit much lower Task CR values, indicating their limited generalization ability.

The superior generalization performance of our EBC-LLMAgent can be attributed to the Behavior Cloning Chain Fusion module, which enables the agent to learn from multiple demonstrations and merge the learned behaviors into a cohesive and flexible interaction model. By dynamically invoking and combining learned functions based on the recognized task requirements, the agent can adapt to new scenarios efficiently. The Behavior Cloning Chain Fusion module allows the agent to leverage the knowledge gained from diverse demonstrations, enabling it to handle a wide range of tasks and exhibit robust and flexible behavior.
The cross-type experiments demonstrate the agent's ability to generalize across different task types within the same application. This generalization capability is crucial for handling the diverse range of user preferences and interactions within a single application. On the other hand, the cross-scene experiments showcase the agent's ability to adapt to different contexts or scenes within an application, ensuring smooth navigation and task completion across varying user interfaces and workflows.

\subsection*{Q5: Explainability Analysis}

To evaluate the explainability of our EBC-LLMAgent, we conducted a comprehensive experiment to assess the quality, usefulness, and human-interpretability of the explanations generated by the agent for its actions. This experiment addresses the ``Explainable'' aspect of our method, demonstrating how our approach not only performs tasks effectively but also provides transparent reasoning for its decision-making process.

We compared EBC-LLMAgent with GPT-4v and AppAgent on three key metrics: Explanation Coherence (EC), which measures how well the explanation aligns with the action taken; Human Alignment (HA), which measures how well humans understand and agree with the explanation; and Explanation Granularity (EG), which assesses the level of detail provided in the explanation. All metrics were scored on a scale of 1 to 5.

Our experimental setup involved 100 diverse actions across the five applications (McDonald's, Starbucks, Booking, YouTube, and WhatsApp), covering various task complexities. A panel of 20 human evaluators, including both AI experts and non-experts, rated the explanations for these actions. For each action, we presented the evaluators with the initial UI state, the action taken by the agent, the resulting UI state, and the explanation provided by the agent. To ensure consistency, we provided detailed rubrics for each metric and conducted a training session with the evaluators before the actual assessment.

\begin{table}[h]
\centering
\caption{Explainability analysis results (average scores across all applications, i.e., McDonald's, Starbucks, Booking, YouTube, and WhatsApp).}
\label{tab:explainability}
\begin{tabular}{lccc}
\hline
Model & EC &  HA & EG \\
\hline
GPT-4v & 3.4 ± 0.3 & 3.2 ± 0.2 & 2.9 ± 0.4 \\
AppAgent & 3.8 ± 0.2 & 3.6 ± 0.3 & 3.5 ± 0.3 \\
EBC-LLMAgent & \textbf{4.5 ± 0.2} & \textbf{4.3 ± 0.2} & \textbf{4.4 ± 0.2} \\
\hline
\end{tabular}
\end{table}

Table \ref{tab:explainability} demonstrates that EBC-LLMAgent consistently outperforms both GPT-4v and AppAgent across all three explainability metrics. EBC-LLMAgent achieved a high EC score of 4.5, indicating that its explanations closely align with the actions taken. This can be attributed to the Demonstration Encoding module, which captures rich semantic information about UI elements and user interactions. With an HA score of 4.3, EBC-LLMAgent's explanations were found to be highly understandable and agreeable by human evaluators, suggesting that our approach bridges the gap between AI decision-making and human interpretability. The EG score of 4.4 demonstrates EBC-LLMAgent's ability to provide detailed explanations without overwhelming users with unnecessary information. This balance is achieved through the Code Generation module, which produces modular and parameterized code snippets with appropriate levels of abstraction.

To illustrate the quality of explanations provided by EBC-LLMAgent, consider the following example from a Starbucks ordering task:

\begin{quote}
User Request: ``Order a Venti Iced Caramel Macchiato with extra shot and light ice.''

EBC-LLMAgent Explanation:
``1. I selected `Caramel Macchiato' from the menu because it matches the base drink in the user's request.
2. I chose `Venti' size as specified by the user.
3. For customization, I selected `Iced' and `Light Ice' options to fulfill the `Iced' and `light ice' requirements.
4. I added an extra shot of espresso by increasing the shot count, addressing the `extra shot' request.
5. Finally, I clicked `Add to Order' to complete the customization process.''
\end{quote}

Each action is driven by specific elements of the user's request, ensuring all requirements are met while navigating the app's interface efficiently. This explanation demonstrates how EBC-LLMAgent breaks down the task into logical steps, maps user requirements to specific UI interactions, provides reasoning for each action taken, and maintains a clear connection between the user's intent and the agent's behavior.

The superior explainability of EBC-LLMAgent can be attributed to several factors. The Demonstration Encoding module captures rich semantic information during the encoding process, enabling the agent to develop a nuanced understanding of UI elements and their relationships. This allows for context-aware explanations that align closely with human perception of the interface. The Code Generation module produces modular and parameterized code, allowing for granular explanations that can be tailored to different levels of technical understanding. This flexibility contributes to the high Human Alignment scores. The UI Mapping component ensures accurate correspondence between generated code and UI elements, grounding explanations in the actual app interface and enhancing coherence and interpretability. Finally, the Behavior Cloning Chain Fusion technique allows the agent to combine learned behaviors in novel ways, enabling it to explain complex sequences of actions by referencing familiar patterns from its training demonstrations.

\section{Limitations and Future Work}

Although our approach demonstrates strong generalization capabilities, there may be cases where the agent encounters completely novel or ambiguous scenarios that differ significantly from the learned demonstrations. In such cases, the agent may require additional guidance or intervention from the user. One potential direction for future work is to incorporate active learning techniques, where the agent proactively seeks user feedback or clarification when faced with uncertainty, allowing it to continually expand its knowledge and adapt to new situations. In addition, scalability is another aspect that warrants further investigation. Our current experiments focused on a limited set of mobile applications, but real-world users interact with a wide variety of apps across different domains. Future research could explore techniques for efficiently scaling our approach to handle a larger number of apps and adapt to the ever-evolving landscape of mobile app interfaces and functionalities. Lastly, while our approach aims to automate mobile app interactions, it is important to consider the potential impact on user privacy and security. Future work should investigate techniques for ensuring the safe and responsible use of such agents, including mechanisms for user control, data protection, and secure communication between the agent and the mobile apps.

Despite these limitations, our work has significant broader implications in various domains. EBC-LLMAgent has the potential to revolutionize the way users interact with mobile apps, enabling seamless and effortless task completion. By automating repetitive and time-consuming tasks, our approach can significantly enhance user productivity and efficiency. Beyond mobile app automation, the principles and techniques introduced in our work can be adapted and applied to other domains, such as web automation, robotic process automation, and intelligent virtual assistants.

\section{Ethical Considerations}

The development and deployment of EBC-LLMAgent for autonomous mobile app interaction raise important ethical considerations that warrant discussion. Our work, while advancing the field of AI-driven app automation, also brings to light several ethical implications that need to be carefully considered.

Firstly, the privacy and data protection aspects of our system are paramount. EBC-LLMAgent interacts with mobile applications that often contain sensitive user data. While our experiments focused on task completion and performance metrics, it is crucial to acknowledge the potential privacy risks involved in autonomous app interaction. Future implementations of such systems must prioritize robust data protection measures to safeguard user information.

Transparency and explainability form core strengths of our approach, as demonstrated by the high scores in Explanation Coherence (EC) and Human Alignment (HA) in our explainability analysis. The ability of EBC-LLMAgent to provide clear, interpretable explanations for its actions enhances user trust and understanding. This aligns with the growing demand for transparent AI systems and addresses concerns about the ``black box’’ nature of some AI technologies.

The generalization capabilities of EBC-LLMAgent, as shown in our cross-type and cross-scene experiments, raise questions about the boundaries of AI autonomy. While our system demonstrates impressive adaptability across different app types and scenarios, it is important to consider the ethical implications of AI systems that can navigate diverse digital environments with minimal human intervention. The balance between automation and user control needs careful consideration to ensure that users maintain agency over their digital interactions.

Our experiments across various mobile applications (McDonald's, Starbucks, Booking, YouTube, and WhatsApp) highlight the potential for EBC-LLMAgent to interact with a wide range of services. This versatility, while technologically impressive, raises questions about the broader societal impact of widespread adoption of such AI agents. Considerations include potential changes in user behavior, the impact on digital literacy, and the implications for app developers and service providers.

The high performance of EBC-LLMAgent in task completion, especially for complex tasks, underscores the need for clear accountability frameworks. As AI agents become more capable of executing complex sequences of actions autonomously, establishing clear lines of responsibility for the outcomes of these actions becomes crucial. This is particularly important in scenarios where the agent's actions might have financial, legal, or personal consequences for the user.

Lastly, the integration of large language models (LLMs) in our approach raises ethical considerations related to the biases and limitations inherent in these models. While our work focused on leveraging LLMs for code generation and task execution, it is important to acknowledge that these models can perpetuate biases present in their training data. Ongoing efforts to address and mitigate these biases are essential for the responsible development of AI systems like EBC-LLMAgent.

In summary, while our work on EBC-LLMAgent represents a significant advancement in autonomous mobile app interaction, it also highlights the need for ongoing ethical consideration and responsible development practices in AI. Balancing the benefits of automation with user privacy, autonomy, and societal implications remains a critical challenge as we continue to push the boundaries of AI capabilities in everyday digital interactions.

\section{Conclusion}
In this study, we introduced Explainable Behavior Cloning LLM Agent (EBC-LLMAgent), a novel approach that combines LLMs with behavior cloning by learning demonstrations to create explainable agents for autonomous mobile app interaction. Our methodology enables LLM agents to learn from user demonstrations, generalize to unseen tasks, and provide transparent explanations of their actions. Our approach opens up new possibilities for creating trustworthy and adaptable agents that can automate complex tasks, reduce human effort, and enhance user experiences across a wide range of mobile applications.

\bibliographystyle{unsrt}
\bibliography{ebc_llm_agent}

\newpage
\appendix

\section{Appendix}
\subsection{Demonstration Encoding Sample}
\begin{figure}[h]
    \centering
    \includegraphics[width=1.0\columnwidth]{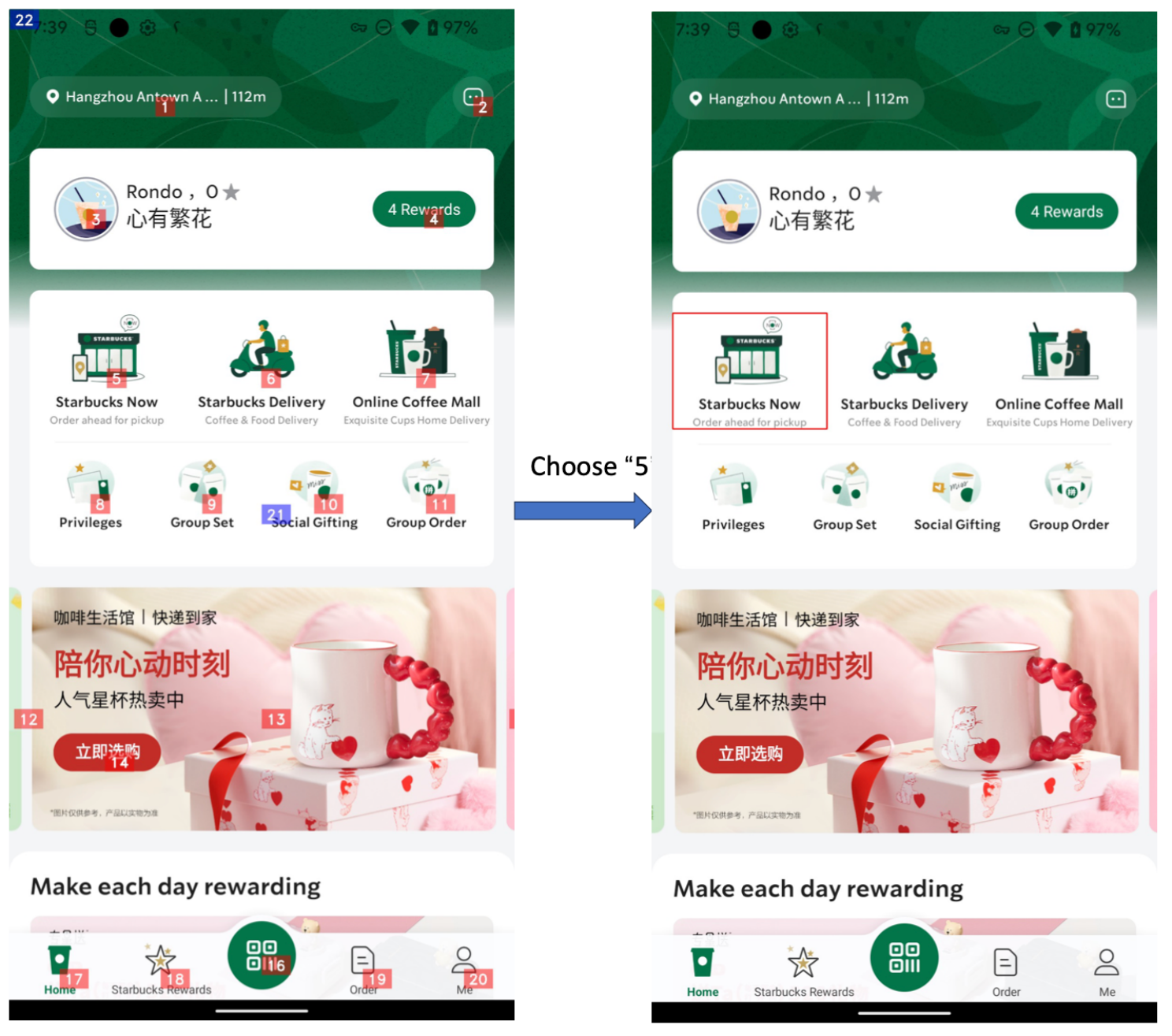}
    \caption{The example of Demonstration Encoding.}
    \label{figure:starbucks_demo}    
\end{figure}

\begin{figure}[h]
    \centering
    \includegraphics[width=1.2\columnwidth]{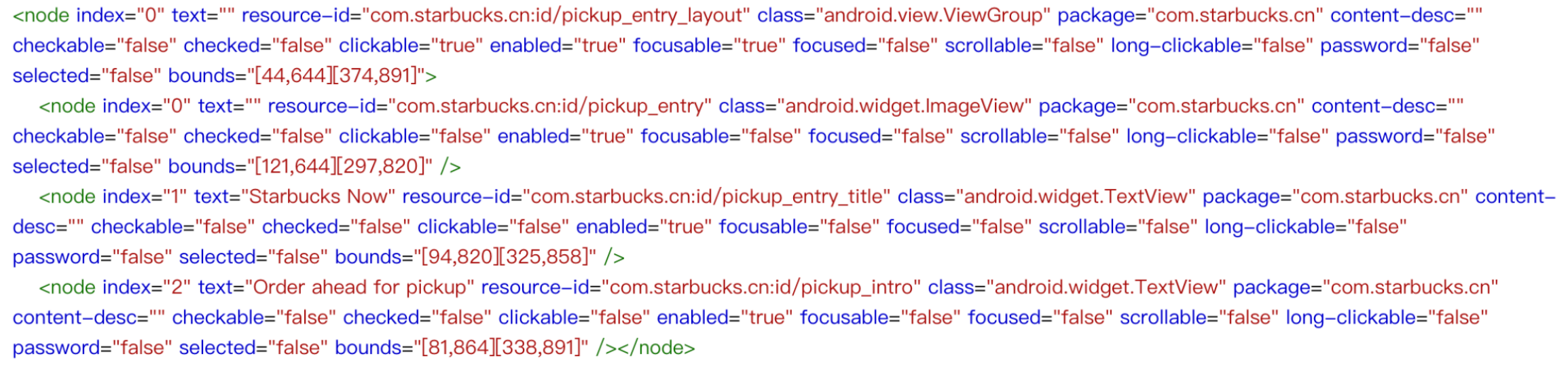}
    \caption{The XML fragment.}
    \label{figure:starbucks_xml}    
\end{figure}

As demonstrated in Table~\ref{figure:starbucks_demo}, by analyzing the original XML file, we are able to mark all clickable elements. The user selected the element of id 5, for which we have annotated the corresponding bounds. Table~\ref{figure:starbucks_xml} presents the XML fragment associated with the element of id 5, highlighting crucial attributes including text, resource\_id, and bounds. Of course, there will also be situations where both text and resource\_id are empty. In such cases, we will rely on multimodal models to generate related text.

\begin{figure}[h]
    \centering
    \includegraphics[width=0.9\columnwidth]{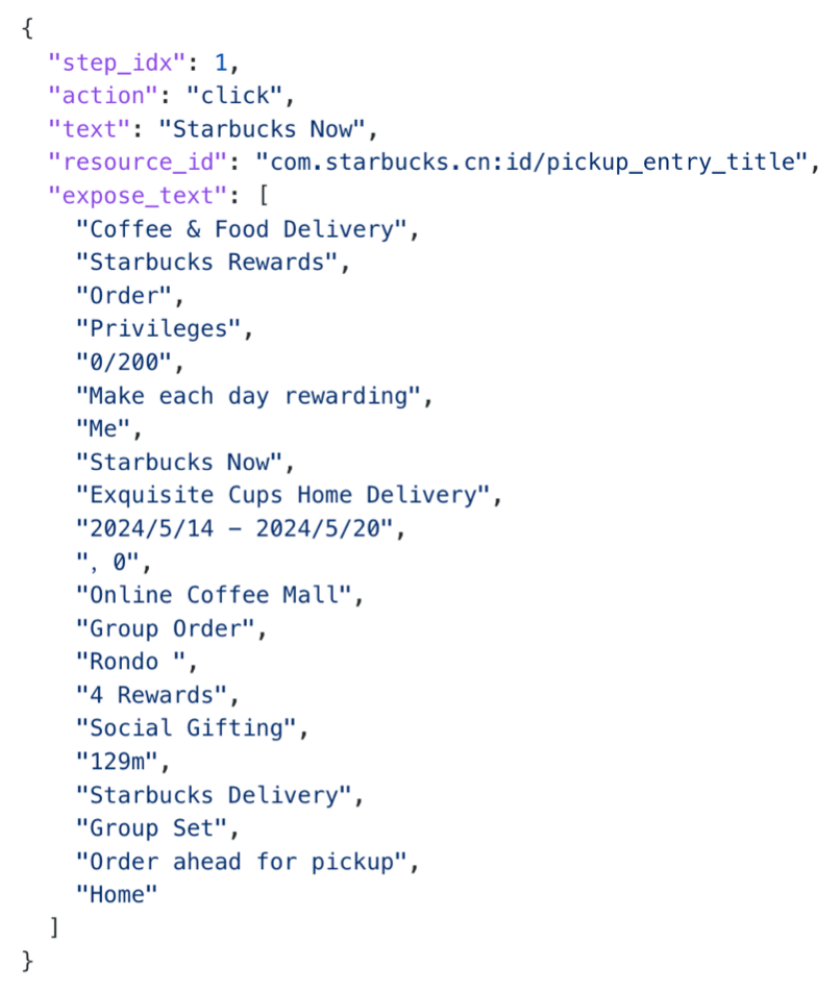}
    \caption{The operation result of Demonstration Encoding}
    \label{figure:operation_step1}    
\end{figure}
Figure~\ref{figure:operation_step1} illustrates the operations demonstrated by users in Figures~\ref{figure:starbucks_demo} and Figure~\ref{figure:starbucks_xml}. When demonstrating a specific task, there are many such steps involved; we compile these steps into a single sequence, and then employ a Large Language Model (LLM) to generate code.

\subsection{Details of Code Generation.}
Given that the code we create is meant to function in an actual environment, we must predefined a set of functions to guarantee the controllability of the end code.
In fact, when generating code, in addition to utilizing our predefined functions, we also require it to possess a certain level of logical structure. For instance, employing 'if' statements for conditional checks and optimizing repetitive steps through loops.
We provide these interfaces for code generation. Including clickAndGetExpose, type, scrollAndGetExpose, and enter. Below are the specific definitions and descriptions of these interfaces.

\begin{figure}[h]
    \centering
    \includegraphics[width=1.2\columnwidth]{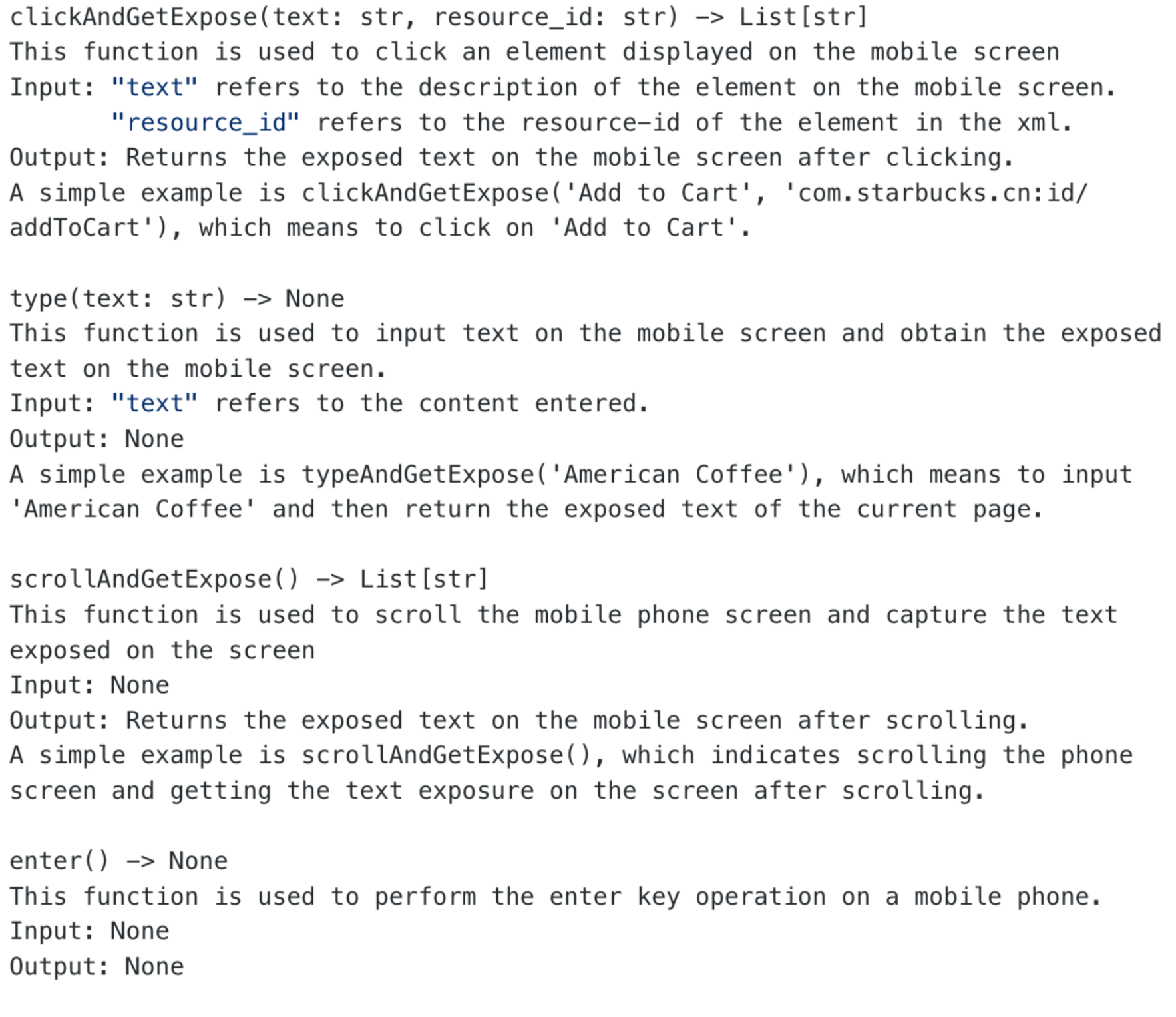}
    \label{figure:api}    
\end{figure}

\begin{figure}[h]
    \centering
    \includegraphics[width=1.2\columnwidth]{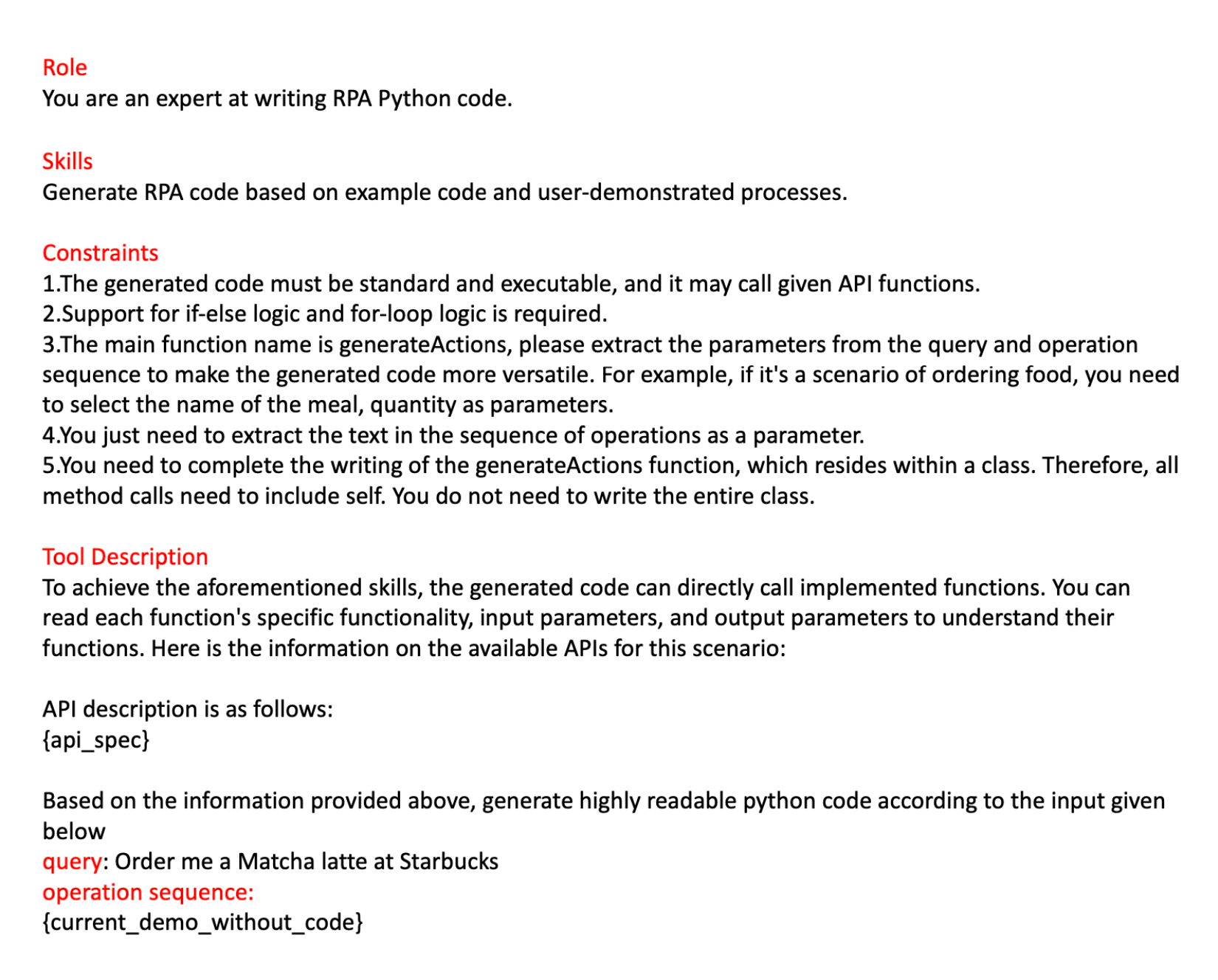}
    \caption{The prompt of code generation.}
    \label{figure:prompt}    
\end{figure}

Figure~\ref{figure:prompt} shows the prompt necessary for our code generation module. We organize it according to the structure of Role, Skills, Constraints, Tool Description, and Operation Sequence, and it has undergone extensive testing in various LLMs.
It should be noted that \{api\_spec\} represents the definition of the functions we provide. \{current\_demo\_with\_code\} denotes the query we demonstrate and the complete operation sequence.

\begin{figure}[h]
    \centering
    \includegraphics[width=1.2\columnwidth]{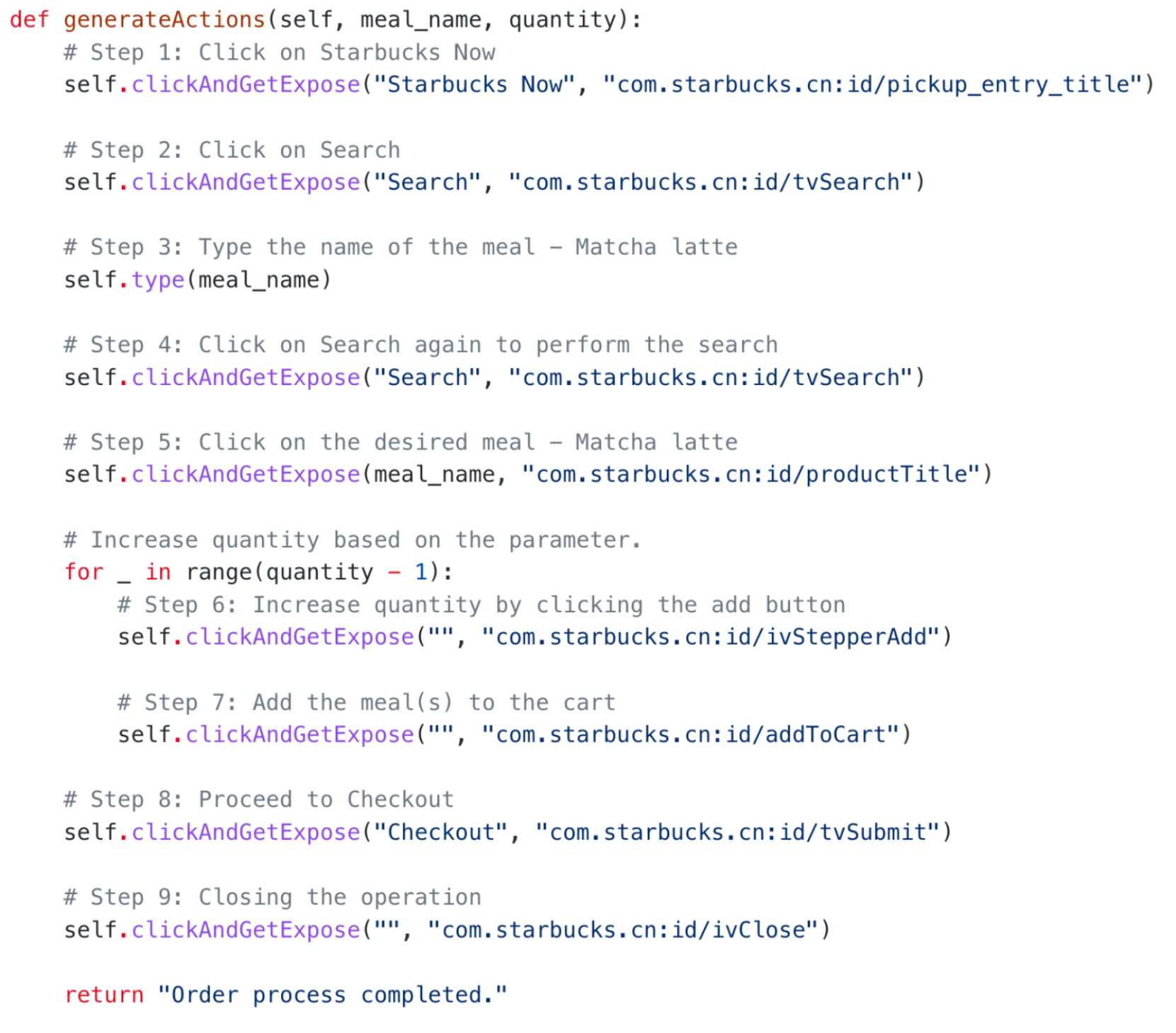}
    \caption{The generate target function.}
    \label{figure:generateActions}    
\end{figure}
Table~\ref{figure:generateActions} presents a snippet of code we generated, suitable for selecting meal items and their quantities at Starbucks. Each step includes explanatory comments, enhancing the readability of the code. Of course, this code requires running within our template.py to function properly. For more details, refer to the code section in the supplementary material.

\end{document}